\documentclass[10pt,twocolumn,letterpaper]{article}

\usepackage{microtype}
\usepackage{graphicx}
\usepackage{svg}
\usepackage{subfigure}
\usepackage{booktabs} 
\usepackage{graphicx}

\usepackage{authblk}

\definecolor{cvprblue}{rgb}{0.21,0.49,0.74}
\usepackage[pagebackref,breaklinks,colorlinks,citecolor=cvprblue]{hyperref}

\usepackage{cvpr}    

\usepackage{amsmath}
\usepackage{amssymb}
\usepackage{mathtools}
\usepackage{amsthm}
\usepackage{float}
\usepackage{amsfonts} 
\usepackage{bbm}      
\usepackage{bm}       
\usepackage{mathrsfs} 
\usepackage{caption}
\usepackage{multirow}
\usepackage{enumitem}
\usepackage[capitalize,noabbrev]{cleveref}
\theoremstyle{plain}

\theoremstyle{definition}

\theoremstyle{remark}

\usepackage[textsize=tiny]{todonotes}

\def\paperID{20} 
\def\confName{CVPR}
\def\confYear{2024}

\title{Situation Monitor: Diversity-Driven Zero-Shot Out-of-Distribution Detection using Budding Ensemble Architecture for Object Detection}

\author[1, 2]{Syed Sha Qutub}
\author[1]{Michael Paulitsch}
\author[1]{Kay-Ulrich Scholl}
\author[1]{Neslihan Kose Cihangir}
\author[1]{Korbinian Hagn}
\author[1]{Fabian Oboril}
\author[2]{Gereon Hinz}
\author[2]{Alois Knoll}

\affil[1]{Intel Labs, Munich, Germany}
\affil[2]{Technical University of Munich, Munich, Germany}

\begin{document}
\maketitle

\begin{abstract}

We introduce Situation Monitor, a novel zero-shot Out-of-Distribution (OOD) detection approach for transformer-based object detection models to enhance reliability in safety-critical machine learning applications such as autonomous driving. The Situation Monitor utilizes the Diversity-based Budding Ensemble Architecture (DBEA) and increases the OOD performance by integrating a diversity loss into the training process on top of the budding ensemble architecture, detecting Far-OOD samples and minimizing false positives on Near-OOD samples.
Moreover, utilizing the resulting DBEA increases the model's OOD performance and improves the calibration of confidence scores, particularly concerning the intersection over union of the detected objects. 
The DBEA model achieves these advancements with a 14\% reduction in trainable parameters compared to the vanilla model. This signifies a substantial improvement in efficiency without compromising the model's ability to detect OOD instances and calibrate the confidence scores accurately. 

\end{abstract}

\section{Introduction}
In machine learning, models must exhibit effective generalization capabilities that require adaptability beyond their training data.
This adaptability is crucial for ensuring the effective performance of models in real-life situations populated with diverse objects. 
In real-world applications, particularly in safety-critical scenarios such as autonomous driving or medical diagnosis, the capability to detect OOD instances is decisive for ensuring the robust performance of machine learning models. 
OOD instances refer to situations where the model encounters data patterns or objects that differ significantly from what it was exposed to during training. Detecting OOD instances becomes particularly challenging when models are expected to generalize effectively across diverse and unpredictable situations \cite{yang2021generalized}. This adaptability is essential for ensuring the reliability and safety of machine learning models in dynamic and complex environments.

\begin{figure}[t!]
\centering
    \includegraphics[width=\linewidth]{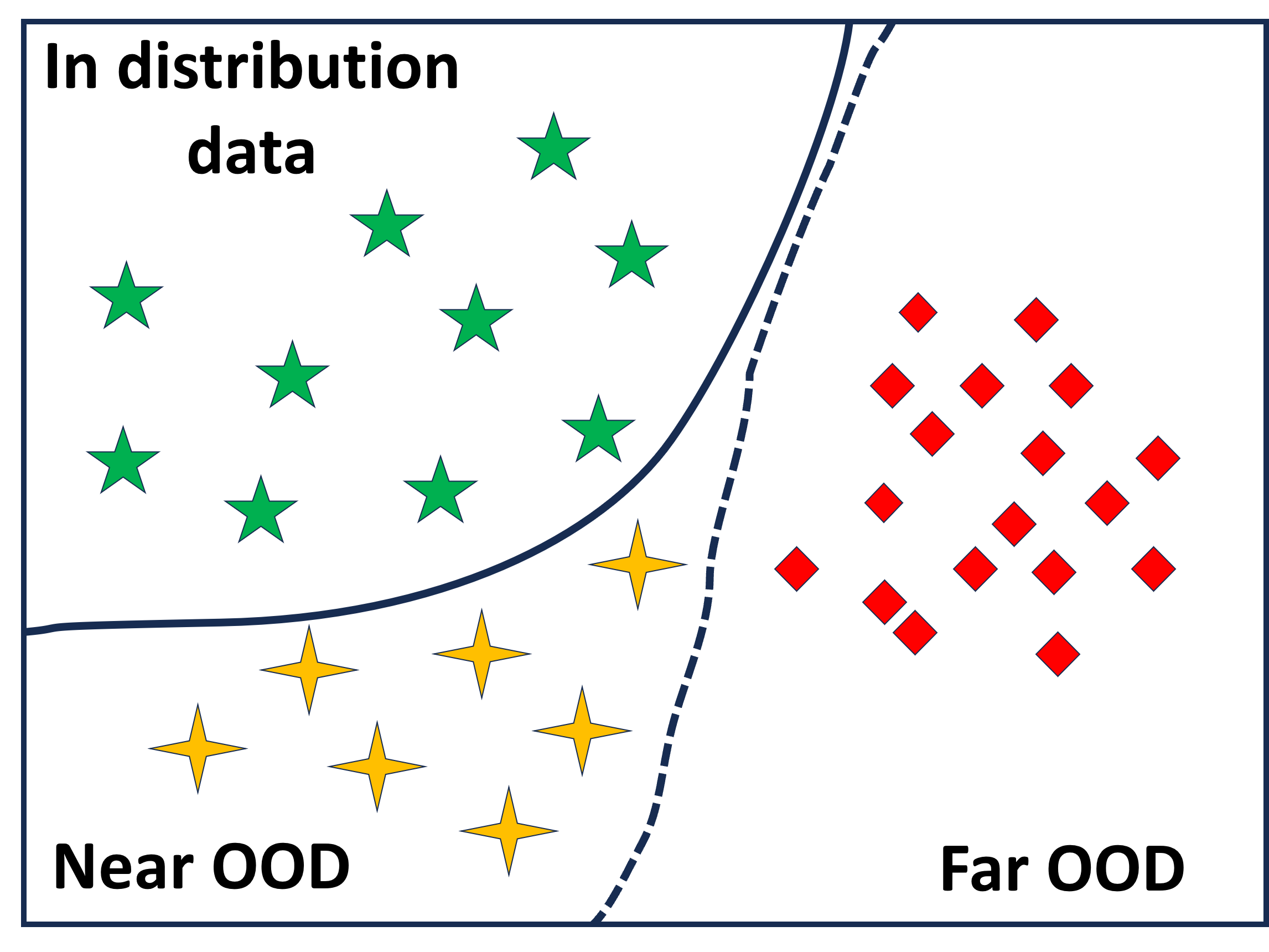}
    \caption{Out-of-Distribution definition, Dotted line represents the decision boundary of an OOD detection model that generalizes effectively.}
    \label{fig:OOD definition}
\end{figure}

To address this classification challenge, this study explores the two types of OOD conditions, as depicted in Fig. \ref{fig:OOD definition}: Near-OOD and Far-OOD \cite{fort2021exploring}. In the context of a Near-OOD dataset, there is a notable resemblance in features and characteristics to the training dataset, also referred to as in-distribution (IN) data. The Near-OOD dataset may resemble datasets obtained from diverse acquisition sensors, e.g., when evaluating a model trained with the autonomous driving dataset KITTI \cite{geiger2012we}, other autonomous driving datasets like BDD100K \cite{yu2020bdd100k}, Cityscapes \cite{cordts2016cityscapes} (2D bounding boxes from CityPersons \cite{zhang2017citypersons}) or Lyft \cite{WovenPlanetHoldingsInc} (2D bounding boxes computed from 3D bounding boxes) can be categorized as Near-OOD. The Far-OOD dataset introduces a different paradigm, where the dataset is entirely dissimilar to the In-Distribution (IN) dataset, surpassing the characteristics of Near-OOD conditions. Given the previous example, the CoCo \cite{lin2014microsoft} dataset would suffice for this Far-OOD condition.
The OOD module integrated within the Deep Neural Networks (DNNs) for a specific application should avoid miss-classifying Near-OOD instances as OOD while correctly flagging instances significantly different from the training dataset, i.e., Far-OOD, as the OOD cases.

With the transition from traditional Convolution Neural Networks (CNNs) to transformer-based models due to their remarkable ability to capture long-range dependencies and contextual information \cite{raghu2021vision}, a shift and redesign of OOD methods designed initially for CNNs is in progress.

Therefore, we showcase the OOD detection performance of our proposed Situation Monitor derived by leveraging the deviations of the ensemble predictions of the DBEA model integrated into the DINO-DETR \cite{zhang2022dino} transformer-based vision model. DBEA is derived from the budding ensemble architecture proposed by \cite{qutub2022bea}. The tandem loss function is hereby extended to incorporate a diversity loss function. Given that vision-based transformers commonly feature encoder-decoder structures. As a final stage of fully connected layers, it is generally applicable to integrate the Situation Monitor into various other vision transformer models without loss of generality.

In this work, we propose to:
\begin{itemize}[noitemsep,nolistsep]
\item Create the DBEA model by incorporating diversity-based loss into the training process of BEA.
\item Define the Situation Monitor to detect Far-OOD samples and suppress false positives on Near-OOD samples.
\item Minimize the overhead of the transformer model with the Situation Monitor compared to baseline models.
\end{itemize}
Through comprehensive ablation study experiments, we evaluate the performance of DBEA in comparison to multiple sample-free OOD detection baselines. These baselines were specifically trained on well-established datasets such as KITTI and BDD100K, serving as benchmarks for our analysis. 
\section{Related Work}

Researchers are continuously exploring innovative methodologies and refining existing techniques to bolster the capabilities of object detectors, particularly in handling out-of-distribution situations. Classifying Deep Neural Networks (DNNs) into deterministic and probabilistic networks provides a foundational understanding.

\textbf{Deterministic Networks vs Probabilistic Networks}: Deterministic networks in DNNs generate consistent outputs for a given input in a deterministic manner \cite{feng2021review}. However, these networks cannot model prediction uncertainty. As a result, confidence scores associated with their predictions become crucial for measuring uncertainty, serving as valuable indicators for OOD detection \cite{berger2021confidence}. In contrast, probabilistic DNNs explicitly model uncertainty in their predictions by outputting probability distributions over possible outcomes \cite{feng2021review}. This explicit modelling benefits OOD detection across a wide range of applications \cite{ahuja2019probabilistic}. 
While OOD detection using probabilistic networks can be computationally demanding, researchers have devised various techniques for deterministic networks, primarily focusing on post-hoc methods \cite{wilson2023safe, kuleshov2018accurate, liang2017enhancing}.

\textbf{Unified Frameworks for OOD Detection}: In addition to post-hoc methods, a few unified frameworks incorporate OOD detection seamlessly into the primary task of the DNN. These frameworks utilize zero-shot or few-shot learning strategies to improve OOD detection capabilities \cite{chen2020boundary, esmaeilpour2022zero}. 

The findings from \cite{kornblith2019similarity}, indicating that wider networks with similar architecture learn more similar features, are integrated into the subsequent work of the sample-free uncertainty estimation method BEA \cite{qutub2022bea}. BEA notably exhibited significant improvements in OOD detection, primarily focusing on anchor-based CNN models for object detection. It is worth noting that most of the related research is connected to CNN-based DNNs, and there is limited exploration into adapting and extending these techniques for transformer-based object detection models.

\textbf{Sample-Free Uncertainty Estimation Method}: Notably, BEA \cite{qutub2022bea}, a sample-free uncertainty estimation method, demonstrated considerable performance enhancements in OOD detection. This method primarily focused on anchor-based CNN models for object detection. Similarly, Gaussian-Yolov3 \cite{choi2019gaussian} introduced Gaussian parameters to exploit variances and reduce false positives by calibrating them to the IoU.

In summary, researchers are continually exploring new methodologies and refining existing techniques to enhance the capability of object detectors in handling out-of-distribution situations. This includes methods focused on both deterministic and probabilistic networks, as well as unified frameworks that seamlessly incorporate OOD detection into the primary task of the DNN. Additionally, there is potential for adapting and extending these techniques for transformer-based object detection models. However, most of this research is associated with CNN-based DNNs, and there is limited exploration into adapting and extending these techniques for transformer-based object detection models.


\section{Problem Statement}
\label{sec:ood_definition_problem_statement}


\begin{figure}
\centering
    \includegraphics[width=\columnwidth, scale=0.35]{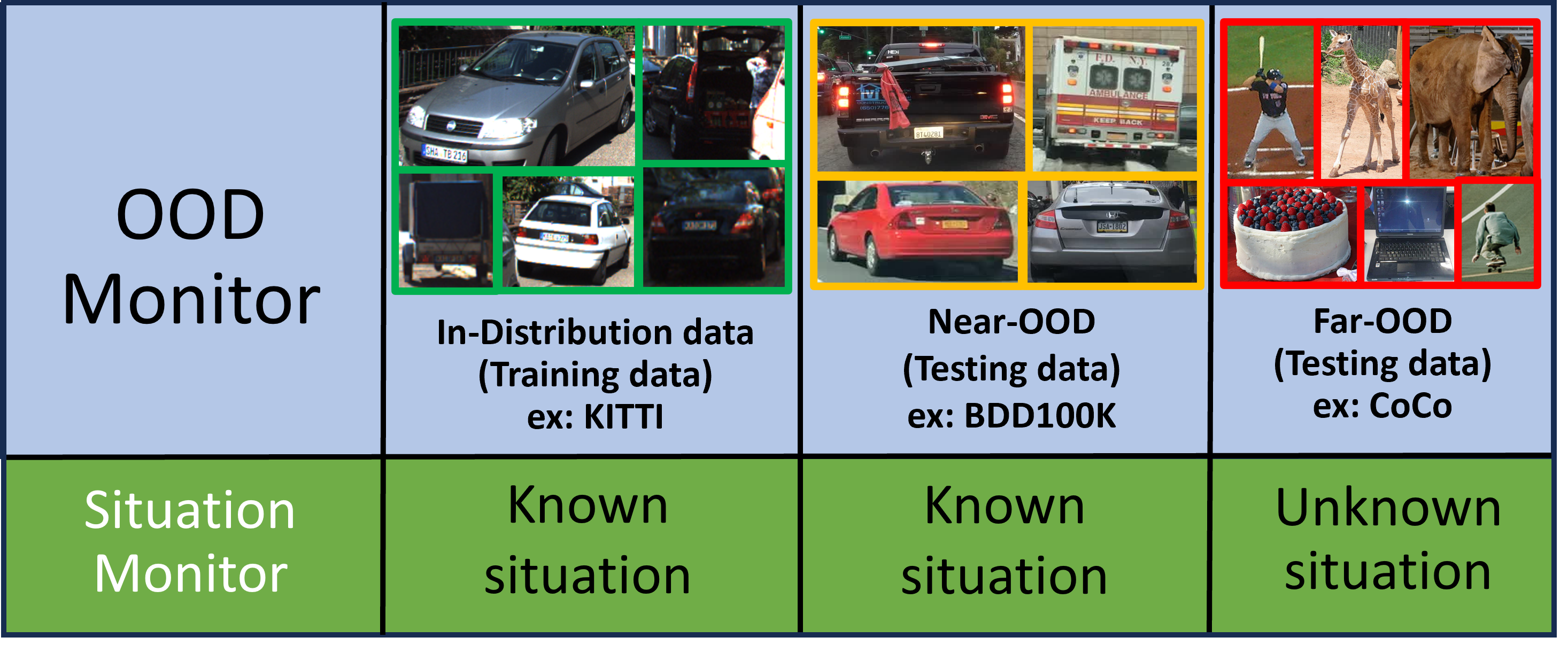}
    \caption{The primary aim of the Situation Monitor is to distinguish between known and unknown situations. For instance, a model trained on datasets such as KITTI for automotive scenarios is categorized as a Near-Out-of-Distribution (Near-OOD) situation. Consequently, encountering an indoor scenario would be labelled as Far-Out-of-Distribution (Far-OOD) situation by the model.}
    \label{fig:OOD Monitors definition}
\end{figure}

\label{subsec:problem_statement}
In the closed-world assumption, the training dataset ($\mathcal{D}$) and testing dataset ($\mathcal{T}$) is from in-distribution dataset $\mathcal{I}$, \textit{i.e.}, $\mathcal{D}$ $\subset$ $\mathcal{I}$. Therefore, the samples from the testing dataset is $\mathcal{S(T)}=\mathcal{S(T^\mathcal{I})}$. However, in open-world settings and practical, real-world scenarios, samples are also drawn from OOD data. Therefore, the OOD samples $\mathcal{O}$ are composed of both Near-OOD ($\mathcal{O^\textit{near}}$) and Far-OOD ($\mathcal{O^\textit{far}}$), \textit{i.e.} $\mathcal{O}$ = $\mathcal{O^\textit{near}}$ + $\mathcal{O^\textit{far}}$. 
Similarly, in the open world setting, the testing data $\mathcal{T}$ consists of known situations and classes originating from Near-OOD $\mathcal{O^\textit{near}}$ and unknown situation and unknown classes to the model originating from Far-OOD $\mathcal{O^\textit{far}}$ \textit{i.e.},
$\mathcal{T} = \mathcal{T^\mathcal{I}} + \mathcal{T}^{\mathcal{O}^{near}} + \mathcal{T}^{\mathcal{O}^{far}}$ and accordingly the samples from the testing data
is $\mathcal{S(T)} = \mathcal{S(T^\mathcal{I})} + \mathcal{S}(\mathcal{T}^{\mathcal{O}^{near}}) + \mathcal{S}(\mathcal{T}^{\mathcal{O}^{far}})$.


A model trained with $\mathcal{D}$ is not only required to detect objects from $\mathcal{S(T^\mathcal{I})}$ but it is also required to generalize well to $\mathcal{S}(\mathcal{T}^{\mathcal{O}^{near}})$. Therefore, a OOD detection module should not flag $\mathcal{S}(\mathcal{T}^{\mathcal{O}^{near}})$ as an OOD, rather only classify the samples from $\mathcal{S}(\mathcal{T}^{\mathcal{O}^{far}})$ as an OOD sample. 

As shown in Fig. \ref{fig:OOD Monitors definition}, the primary purpose of the Situation Monitor is to distinguish between known and unknown situations. Leveraging the remarkable generalization capability of deep learning models \cite{pinto2022impartial}, the monitor adeptly classifies Near-OOD situations as known situations. This classification is grounded in the understanding that Near-OOD instances share significant similarities with the In-Distribution (IN) dataset, aligning with the inherent generalization prowess of deep learning models. Consequently, the Situation Monitor is pivotal in identifying situations based on their familiarity with the model's learned context. By accurately discerning known and unknown situations, the monitor empowers the model to make informed decisions in real-world applications, even when encountering instances beyond its training data. This enhances the model's reliability and performance in diverse and dynamic environments. Overall, the Situation Monitor plays a crucial role in ensuring the effectiveness of deep learning models across various situations.


In summary, our goal is to train a transformer-based object detection model on a training set $\mathcal{D}$ and without introducing the samples from $\mathcal{O}$, the Situation Monitor (OOD detection module) of the model should have ability to classify the samples from $\mathcal{S}(\mathcal{T}^{\mathcal{O}^{far}})$ as an OOD situation and on contrary should not flag the $\mathcal{S}(\mathcal{T}^{\mathcal{O}^{near}})$ samples as an OOD situation.


\section{Our Approach}
\subsection{DBEA: Diversity based Budding Ensemble Architecture}
The recently introduced Budding Ensemble Architecture (BEA) \cite{qutub2022bea} represents a sample-free methodology for detecting OOD instances, showcasing notable performance advancements. Consequently, we employ the Budding Ensemble Architecture approach in the design of our Situation Monitor. 

\begin{figure}
\centering     
\subfigure[Abstract DINO-DETR architecture]{\label{fig:abstract_arch_a}\includegraphics[width=0.91\columnwidth]{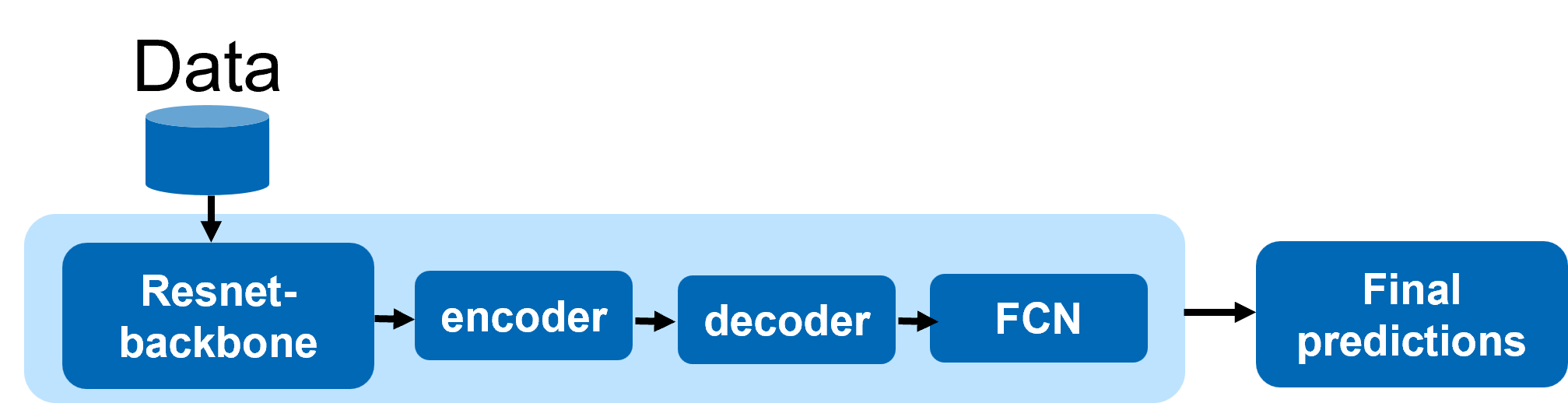}}
\subfigure[Abstract DBEA-DINO-DETR architecture]{\label{fig:abstract_arch_b}\includegraphics[width=0.91\columnwidth]{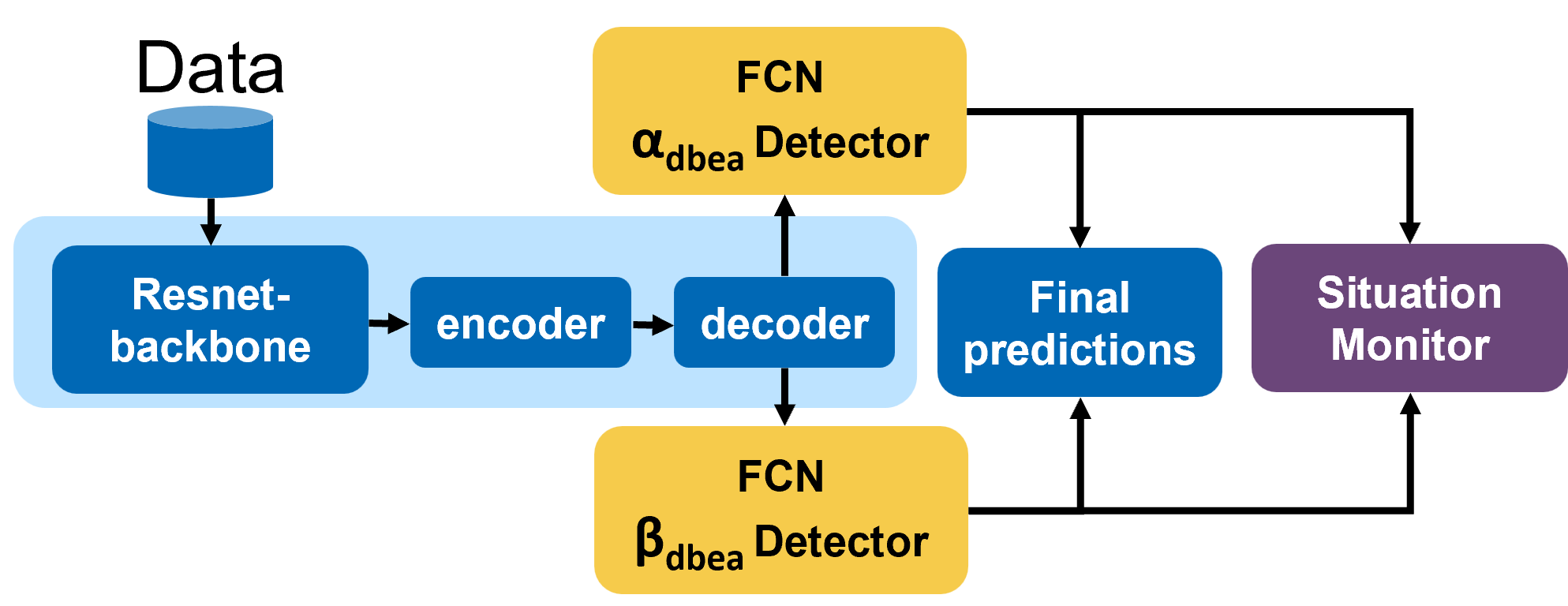}}
\caption{Adaptation of BEA \cite{qutub2022bea} to DINO-DETR \cite{zhang2022dino}.}
\label{fig:abstract_arch}
\end{figure}

Within the Budding Ensemble Architecture (BEA) framework, a unified backbone and duplicated detectors replace the conventional ensemble setup. This alteration enhances confidence score calibration, diminishes uncertainty errors, and introduces an overlooked advantage: superior OOD detection compared to other state-of-the-art sample-free methods. The \textbf{Tandem loss} function ($\mathcal{L_{\mathbf{tandem}}}$) introduced in BEA, devised initially for YOLOV3 \cite{redmon2018yolov3} and SSD \cite{liu2016ssd} was primarily tailored for anchor-based object detection models. Therefore, the $\mathcal{L_{\mathbf{tandem}}}$ function to be applied to the transformer model requires a series of modifications and adaptations to integrate with this novel ensemble approach seamlessly.

\begin{figure}
\centering
    \includegraphics[width=0.65\columnwidth, scale=0.35]{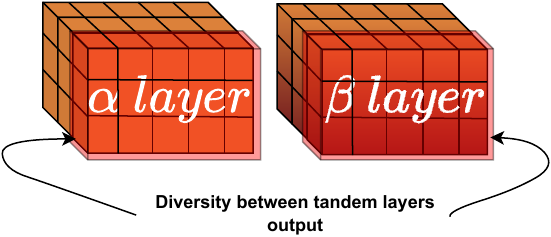}
    \caption{Within the BEA model, tandem detectors undergo further refinement with diversity loss, aiming to enhance the distinction between $\alpha$ and $\beta$ detectors. This leads to the development of Diversity-based BEA.
    }
    \label{fig:diversity_fig}
\end{figure}

\begin{figure*}[h!]
\centering 
    \includegraphics[width=0.95\textwidth]{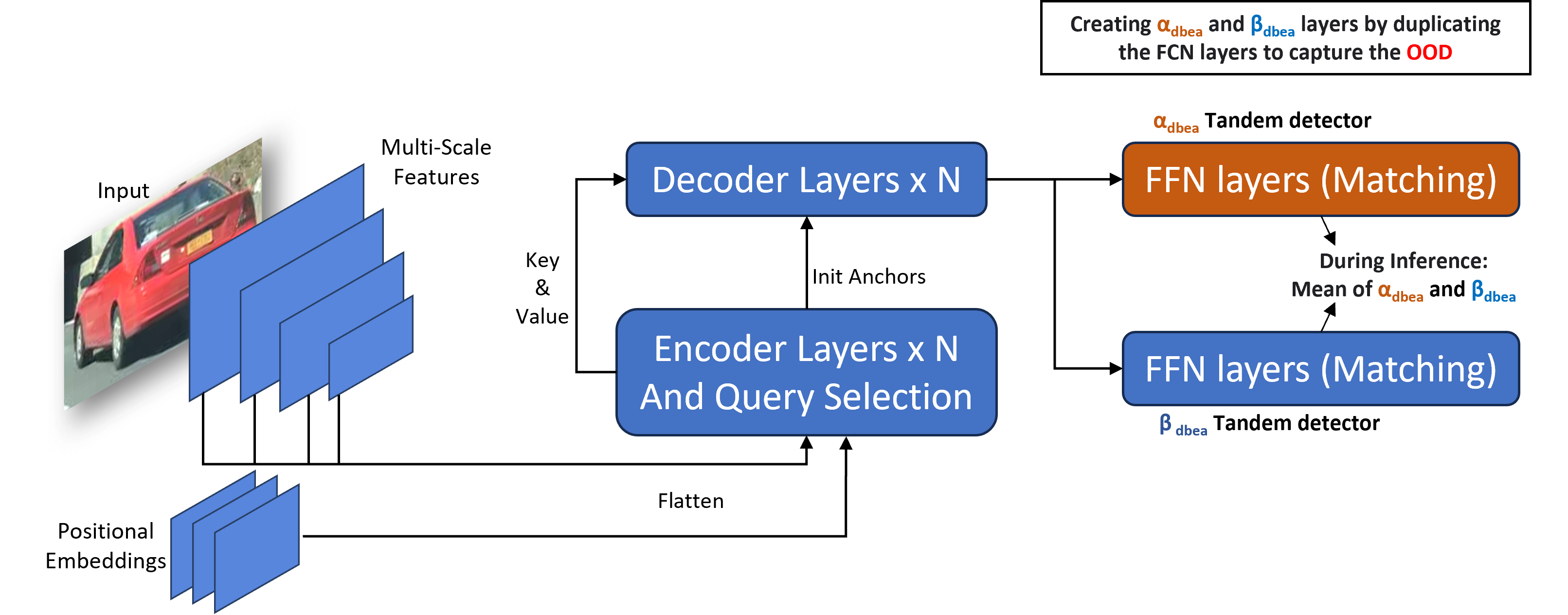}
    \caption{Adapted architecture diagram from DINO-DETR \cite{zhang2022dino} for DBEA-DINO-DETR, illustrating the replication of final layers. For a comprehensive understanding of DINO-DETR, please refer to Figure 2 in \cite{zhang2022dino}. In BEA, it is proposed to duplicate the final layers as $\alpha$ and $\beta$ to create tandem detectors, which are subsequently utilized in the situation monitoring model.}
    \label{fig:BEA_DINO}
\end{figure*}

We base our Situation Monitor design on the ResNet-based DINO-DETR transformer model. It typically comprises a CNN-based backbone (ResNet), self-attention layer-based encoders, and decoders, followed by regression layers for classification and bounding box detection. We leverage the insights from BEA to tailor the DINO-DETR object detection model for our Situation Monitor. Specifically, we introduce \textbf{diversity-based tandem detection layers}, as depicted in Fig. \ref{fig:abstract_arch_b}. In the architectural integration of BEA into DINO-DETR, the 3 layer feed-forward neural network layers (FFN) forming the final regression layers immediately following the decoder are duplicated, resulting in two detectors, $\mathbf{\alpha}$ and $\mathbf{\beta}$ collectively called as tandem detectors. During inference, confidence scores and bounding boxes are computed from the mean of the tandem detectors.

Despite the duplication of layers, as indicated in the BEA paper \cite{qutub2022bea}, no advantages were observed for the Situation Monitor during the training of the DINO-DETR model. This lack of advantage stems from the tandem layers learning similar representations. Therefore, the incorporation of the tandem loss function, $\mathcal{L_{\mathbf{tandem}}}$, alongside the original base loss function, $\mathcal{L_{\mbox{base}}}$, becomes crucial. The adaptation of BEA with the diversity-based $\mathcal{L_{\mathbf{tandem}}}$ function is called Diversity-based Budding Ensemble Architecture (DBEA). This serves a dual purpose: creating an effective Situation Monitor and enhancing the calibration of confidence scores.

The adapted tandem loss is now defined as follows:
\begin{equation}
\label{eq:tandem_aiding_loss}
\begin{aligned}
    \mathfrak{L_{\mbox{ta}}(\phi}) &= \mathbbm{1}_{\textit{i}}^{\mbox{obj}}  \sqrt{\left(\phi^{\alpha}_i - \phi^{\beta}_i\right)^2}, \\ 
    \mathcal{L_{\mbox{ta}}} &=  \mathfrak{L_{\mbox{ta}}}(\boldsymbol{x}) + \mathfrak{L_{\mbox{ta}}}(\boldsymbol{y}) + \mathfrak{L_{\mbox{ta}}}(\boldsymbol{w}) + \mathfrak{L_{\mbox{ta}}}(\boldsymbol{h})
\end{aligned}
\end{equation}

\begin{equation}
\label{eq:tandem_quelling_loss}
\begin{aligned}
    \mathfrak{L_{\mbox{tq}}(\phi}) &= \mathbbm{1}_{\textit{i}}^{\mbox{noobj}}  \frac{1} {\sqrt{\left(\phi^{\alpha}_i - \phi^{\beta}_i\right)^2}}, \\ 
    \mathcal{L_{\mbox{tq}}} &=  \mathfrak{L_{\mbox{tq}}}(\boldsymbol{x}) + \mathfrak{L_{\mbox{tq}}}(\boldsymbol{y}) + \mathfrak{L_{\mbox{tq}}}(\boldsymbol{w}) + \mathfrak{L_{\mbox{tq}}}(\boldsymbol{h})
\end{aligned}
\end{equation}

\begin{equation}
\label{eq:tandem_loss}
\begin{aligned}
    \mathcal{L_{\mbox{tandem}}} &= \lambda_{ta}\mathcal{L_{\mbox{ta}}} + \lambda_{tq}\mathcal{L_{\mbox{tq}}} \\
\end{aligned}
\end{equation}

where $\mathbbm{1}_{\textit{i}}$ denotes whether the object prediction overlaps with the ground-truth.The variables $\boldsymbol{x}$ and $\boldsymbol{y}$ signify the predicted centre points of the bounding box, while $\boldsymbol{w}$ and $\boldsymbol{h}$ represent the predicted height and width of the object, respectively.
The $\mathcal{L_{\mathbf{tandem}}}$ operates on positive predictions and negative predictions, which is possible due to access to ground truth during training. The \textbf{Tandem-Aiding loss} function $\mathcal{L_{\mathbf{ta}}}$ diminishes the errors associated with the positive predictions between $\mathbf{\alpha}$ and $\mathbf{\beta}$ detector. Similarly, the \textbf{Tandem-Quelling loss} function $\mathcal{L_{\mathbf{tq}}}$ amplifies the errors related to negative predictions between $\mathbf{\alpha}$ and $\mathbf{\beta}$ detectors. Therefore promoting agreement and disagreement concerning positive and negative predictions.

Within the BEA paper, the $\mathcal{L_{\mbox{ta}}}$ and $\mathcal{L_{\mbox{tq}}}$ loss functions are applied independently to specific components of both classification and bounding box regression outputs. In the context of the transformer model, the $\mathcal{L_{\mathbf{tandem}}}$ is exclusively applied to the bounding box regression layers as shown in Eq. \eqref{eq:tandem_aiding_loss} and \eqref{eq:tandem_quelling_loss}. This selective application is adopted because introducing this loss to the classification layer decreases the performance of the Situation Monitor.
To support $\mathcal{L_{\mbox{tandem}}}$ loss function for better calibration of the confidence scores, we introduce diversity at classification regression layer output between the tandem layers as shown in Fig. ~\ref{fig:diversity_fig}. To this end, we use similarity score as a diversity measure to ensure that the tandem detectors capture distinct representations from the decoders. Specifically, we use the cosine similarity score to introduce the diversity as depicted in Eq. \eqref{eq:cosin_sim_loss}.

\begin{equation}
\label{eq:cosin_sim_loss}
\begin{aligned}
    Cosine\_Similarity(\alpha, \beta) = \frac{\alpha.\beta}{||\alpha||.||\beta||}; \alpha, \beta \in Z \\
    \mathcal{L_{\mbox{diversity}}}  = \frac{1}{n} \sum_{i=1}^{n}Cosine\_Similarity(\mathfrak{\alpha}_{i}, \beta_{i}) \\
\end{aligned}
\end{equation}
where $n$ represents total number of predictions and $Z$ represents the classification logits denoted as $Z = (z_{1}, z_{2},..z_{k})$, $z_{i}$ is the unnormalized score for class $i$.

Incorporating diversity introduces a dynamic range of classification outputs at the tandem layers. This diversity is instrumental in fostering a broader spectrum of values, thereby providing the tandem loss function with a larger space to operate upon. Specifically, the tandem loss function leverages this diversity to mitigate errors associated with positive predictions effectively. By encouraging divergence in classification values, the model can better discern and refine its understanding of positive instances. Simultaneously, this diversity amplifies errors in the context of negative predictions.
To this end, the diversity and tandem-based loss function is constructed by incorporating both $\mathcal{L_{\mathbf{tandem}}}$ and $\mathcal{L_{\mathbf{diversity}}}$ into the base loss function of the DINO-DETR model, denoted as $\mathcal{L_{\mathbf{base}}}$. This integration is illustrated in Equation \eqref{eq:bea_loss}.  
\begin{equation}
\label{eq:bea_loss}
\begin{aligned}
    \mathcal{L_{\mbox{dbea}}} &= \mathcal{L_{\mbox{base}}} + \mathcal{L_{\mbox{tandem}}} + \lambda_{div}\mathcal{L_{\mbox{diversity}}} \\
\end{aligned}
\end{equation}
The training hyper-parameters of the DBEA-DINO-DETR remain unchanged except the $\lambda$ parameters introduced in Eq. \eqref{eq:tandem_loss} and \eqref{eq:bea_loss}.

\subsection{Situation Monitor}
\label{situation_monitor}
The Situation Monitor is a component of the BEA-DINO-DETR that serves as an Out-of-Distribution (OOD) detection module. It identifies Far-OOD situations by analyzing disparities in the predictions of tandem layer bounding boxes. The transformer model undergoes end-to-end training through a zero-shot approach, which, in this case, means an explicit OOD dataset is not shown during the training process. This methodology distinguishes explicit situations by highlighting the errors in variance between the tandem layer predictions, specifically in Far-OOD situations. The $\mathcal{L_{\mbox{dbea}}}$ loss function, incorporating cosine diversity during training, compels tandem detection layers to consider objects from diverse feature maps and perspectives.

\begin{figure}
\centering
    \includegraphics[width=0.48\textwidth, scale=0.35]{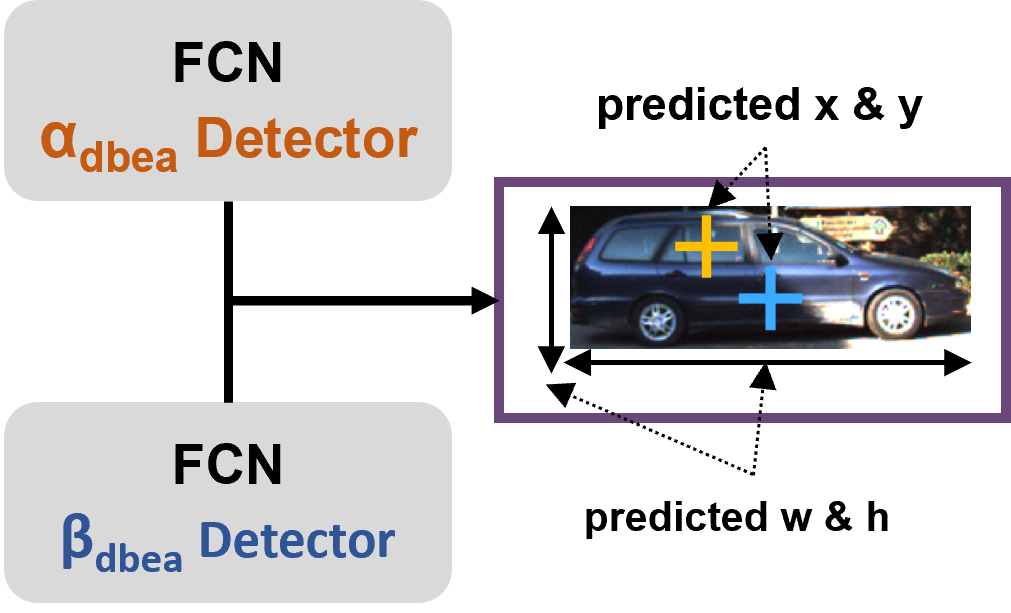}
    \caption{Situation Monitor: The variance between $\alpha_{DBEA}$ and $\beta_{DBEA}$ detector predictions is interpreted as the prediction uncertainty $\mathcal{U}_{SM}$. A high uncertainty means Far-OOD, whereas low uncertainty means Near-OOD or in-distribution.}
    \label{fig:situation_monitor}
\end{figure}

The introduction of diversity loss ($\mathcal{L_{\mbox{diversity}}}$) during training prompts a change in perspective, compelling tandem detection layers to examine objects from diverse feature maps and viewpoints. Driven by the $\mathcal{L_{\mbox{tandem}}}$ loss function, tandem layers generate distinct predictions for bounding box centre points (Fig. \ref{fig:situation_monitor}). 
In instances where the set $\mathcal{T}$ is a subset of $\mathcal{I}$ within the specific category $\mathcal{O^\textit{near}}$, it is observed that the predicted widths and heights tend to be closely aligned with no large variances. Similarly, we should expect large variances when $\mathcal{O^\textit{far}}$ is encountered. Therefore, to capture Far-OOD situations effectively at the image level, a methodology is applied to heighten the variances of predicted centre points ($\boldsymbol{x}$ and $\boldsymbol{y}$) and heights and widths ($\boldsymbol{w}$ and $\boldsymbol{h}$). These bounding box-related attributes are referred to as $b$ in the following equation.

\begin{equation}
\label{eq:OOD_final}
\begin{aligned}
   xy_{var} = \sqrt{\sum_{i \in (x, y)}{(b^{\alpha}_{i} - b^{\beta}_{i}})^2}, \\
   xy_{centered\_var} = xy_{var}*\mu(xy_{var}), \\
   wh_{var} = \sum_{i \in (w, h)}{(b^{\alpha}_{i} - b^{\beta}_{i}})^2, \\
   wh_{centered\_var} = wh_{var}*\mu(wh_{var}) ,\\
   \mathcal{U}_{SM} = \mu(\sqrt{xy_{centered\_var}}*wh_{centered\_var})
\end{aligned}
\end{equation}

To heighten sensitivity to deviations and pinpoint errors at an image-specific level, these variances undergo additional processing as shown in Eq. \eqref{eq:OOD_final}. This involves centring variances by multiplication with their respective means. The final OOD value, denoted as $\mathcal{U}_{SM}$, is computed as the mean of these variances at the image level, as depicted in Eq. \eqref{eq:OOD_final}. This approach effectively addresses global trends across the entire dataset while localizing errors to individual images. By doing so, anomalies unique to each image are identified, enhancing performance in capturing Far-OOD situations.

\section{Evaluation Metrics for Situation Monitor}
\label{sec:metrics}
The impact of adding $\mathcal{L_{\mbox{tandem}}}$ and $\mathcal{L_{\mbox{diversity}}}$ to the $\mathcal{L_{\mbox{base}}}$ loss function is evaluated using \textbf{mean average precision (mAP)} metrics. The calibration is evaluated using \textbf{Pearson correlation (PCorr)} between confidence scores and the intersection over union of objects and the ground truths. Pearson correlation is evaluated on two sets \textbf{PCorr-all} and \textbf{PCorr-tp} which is correlation on all the predictions and separately on true positive samples.
The Situation Monitor in a DBEA model uses $\mathcal{U}_{SM}$ values and it is assessed using four standard metrics to evaluate it's performance in detecting OOD situations. Whereas, the vanilla baseline model's OOD detection is evaluated using their confidence scores. 
In this study, we do not assess the object detection performance (mAP/AP) on Near-OOD datasets. Instead, we simply demonstrate that the Situation Monitor does not identify Near-OOD samples as OOD.
\begin{itemize}[noitemsep,nolistsep]
\item \textbf{AUROC} calculates the area under the receiver operating characteristic curve, a metric utilized to assess the performance of OOD detection. The OOD ($\mathcal{T}(\mathcal{T}^{\mathcal{O}^{far}})$) samples are considered as positive samples. A higher AUROC value indicates superior performance.
\item \textbf{AUPR(In/Out)} is the area under the receiver operating characteristic curve and is a key metric for evaluating OOD detection performance. It assesses how well a model distinguishes positive instances, with $\mathcal{T}(\mathcal{T}^{\mathcal{O}^{far}})$ samples considered as positives. AUPR comprises of AUPR-In and AUPR-Out. It considers the in-distribution ($\mathcal{T}(\mathcal{T}^{\mathcal{I}})$) samples as either positive or negative respectively. A higher AUPR value indicates superior model performance.
\item \textbf{FPR@95} expressed as FPR (false positive rate) at a fixed TPR (true positive rate) point represents the rate of falsely identified positive instances among all negative instances when the true positive rate is held at a specific percentage which in this case at 95\%. The lower value indicates superior performance.
\item \textbf{DE@95} is the detection error at 95\% TPR quantifies the detection error (miss-classification probability) when TPR is set at 95\%. A lower value indicates superior performance.
\end{itemize}


\section{Experiments}
In this section, we conduct an ablation study to analyze the impact of each component. We then compare our method with the previous state-of-the-art sample-free OOD detection approach.
\subsection{Experiment Setup}

The Situation Monitor (SM) is integrated into the DBEA-DINO-DETR object detection model, allowing end-to-end training without freezing the backbone or any particular layer. This ensures tandem detectors learn based on the $\mathcal{L_{\mbox{dbea}}}$ loss function. Evaluations are conducted on the KITTI \cite{geiger2012we} and BDD100K \cite{yu2020bdd100k} datasets, widely used computer vision datasets for autonomous driving situations. 
These datasets are divided into training, validation, and testing sets with ratios of 7.5:1:1.5 and 8.5:0.75:0.75, respectively. Eight out of nine usable classes are evaluated for the KITTI dataset, while all ten classes are considered for the BDD100K dataset. Additionally, CoCo's evaluation dataset is utilized.
A consistent $416\times416$ input image size is used for training across all models.
For YoloV3 \cite{redmon2018yolov3} and SSD \cite{liu2016ssd} models, SGD optimizer is employed with a learning rate of 0.001, momentum of 0.95, and weight decay of 0.001, trained for 300 epochs with batch-size of 28. DINO-DETR-based models utilize 900 queries, generating 900 predictions per image, with the top 100 predictions considered.
ResNet-50-based transformer models use an AdamW optimizer with a weight decay of 0.0001 and a learning rate of 0.0002, trained for 50 epochs with a batch size of 5. Both the baseline and our approach share identical training hyperparameters, ensuring consistency.

\begin{figure}[t]
\centering
    \includegraphics[width=0.95\columnwidth, scale=0.5]{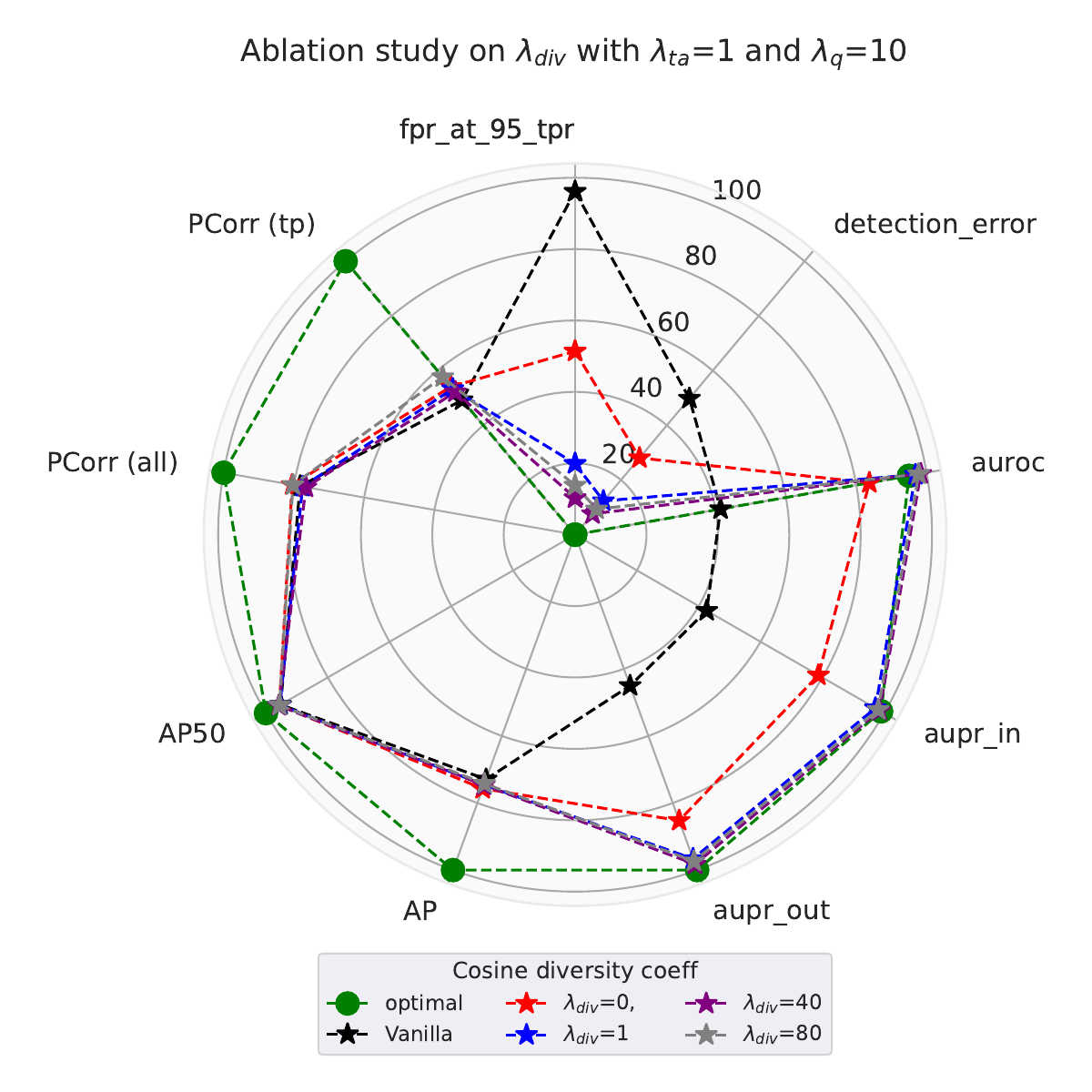}
    \caption{Analysis of the impact of varying parameter $\mathcal{L_{\mbox{div}}}$ in the ablation study.}
    \label{fig:ablation_cosin}
\end{figure}

\subsection{Ablation Study on $\lambda_{div}$,  $\lambda_{ta}$ and $\lambda_{tq}$ on KITTI trained DBEA-DINO-DETR}
In this ablation study section, multiple DBEA-DINO-DETR models are trained on the KITTI dataset wherein the parameters outlined in equations \ref{eq:bea_loss} and \ref{eq:tandem_loss} are varied. The primary objective is systematically showcasing and analysing each component's distinctive impact. In all ablation experiments, the parameter $\lambda_{ta}$ is consistently maintained at either one when necessary or at zero. This is because $\mathcal{L_{\mbox{ta}}}$ is designed to minimize errors in positive predictions, and an increase in the $\lambda_{ta}$ factor beyond $\lambda_{tq}$ adversely affects calibration and the fundamental object detection capability of the model.

\begin{figure}[t]
\centering
    \includegraphics[width=0.95\columnwidth, scale=0.5]{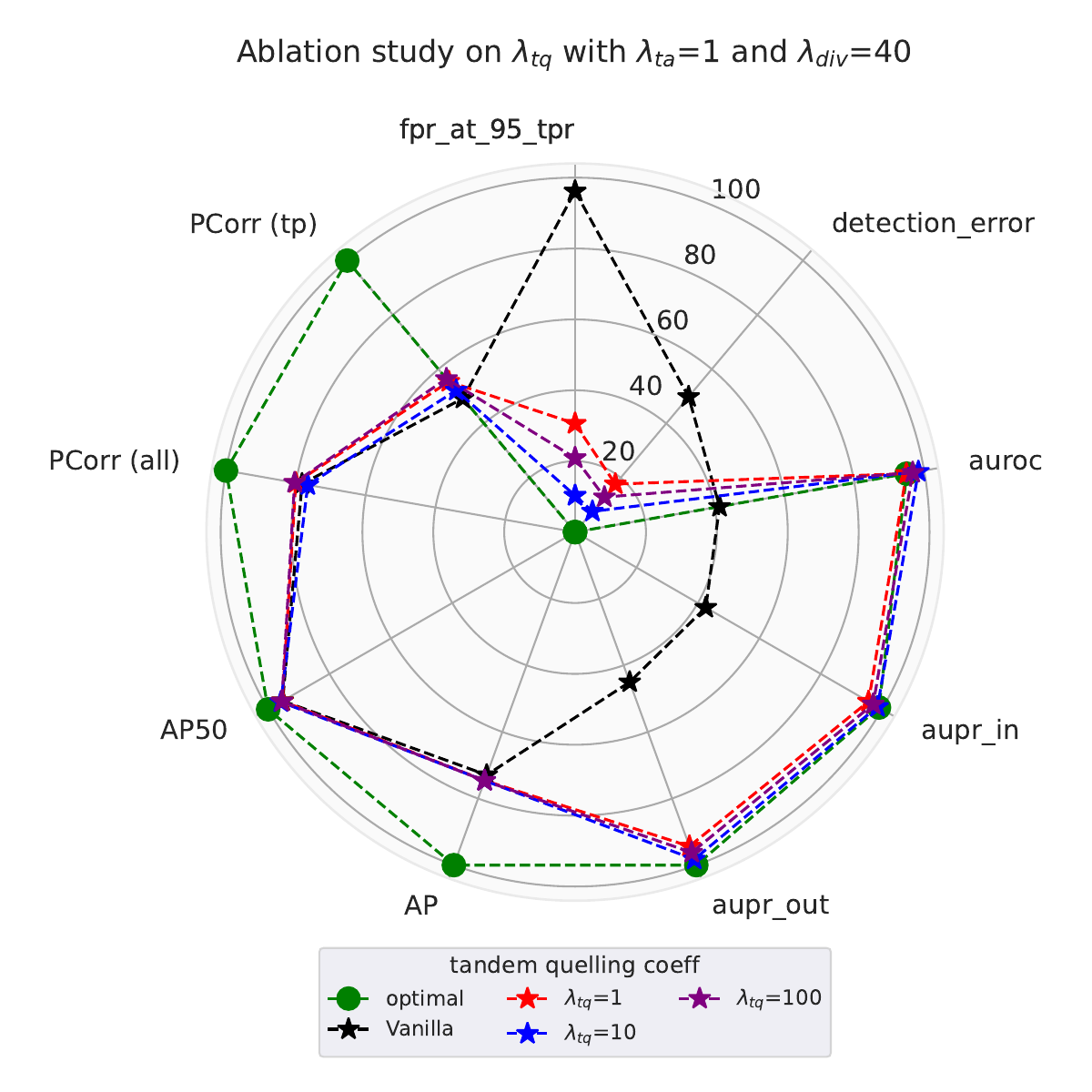}
    \caption{Analysis of the impact of varying parameter $\mathcal{L_{\mbox{tandem}}}$ in the ablation study.}
    \label{fig:ablation_tq}
\end{figure}
\begin{table*}[tbp]
\centering
\captionsetup{justification=centering}
\resizebox{\textwidth}{!}{%
\begin{tabular}{c|c|c|c|c|cccc}
\hline
\multirow{2}{*}{\textbf{\begin{tabular}[c]{@{}c@{}}Model \end{tabular}}} & \multirow{2}{*} {\textbf{\begin{tabular}[c]{@{}c@{}}\textbf{mAP} \\ (\%) $\mathbf{\uparrow}$\end{tabular}}} & \multirow{2}{*}{\textbf{AP50 (\%) $\mathbf{\uparrow}$}} & \multirow{2}{*} {\textbf{\begin{tabular}[c]{@{}c@{}}PCorr(all) \\ (\%) $\mathbf{\uparrow}$\end{tabular}}}  & \multirow{2}{*} {\textbf{\begin{tabular}[c]{@{}c@{}}PCorr(tp) \\ (\%) $\mathbf{\uparrow}$\end{tabular}}} & \multicolumn{4}{c}{\textbf{\begin{tabular}[c]{@{}c@{}}Out-of-distribution detection on COCO dataset \end{tabular}}} \\ 
\cline{6-9}  & & &  &  & \multicolumn{1}{c|}{\textbf{\begin{tabular}[c]{@{}c@{}}AUROC \\(\%) $\mathbf{\uparrow}$ \end{tabular}}} & \multicolumn{1}{c|}{\textbf{\begin{tabular}[c]{@{}c@{}}AUPR (In/Out) \\(\%) $\mathbf{\uparrow}$ \end{tabular}}} & \multicolumn{1}{c|}{\textbf{\begin{tabular}[c]{@{}c@{}}FPR@95 \\(\%) $\mathbf{\downarrow}$ \end{tabular}}} & \multicolumn{1}{c}{\textbf{\begin{tabular}[c]{@{}c@{}}DE@95  \\(\%) $\mathbf{\downarrow}$ \end{tabular}}}\\ \hline \hline 
\multicolumn{2}{l}{\textbf{KITTI trained}}\\ \hline \hline
Vanilla-YOLOv3 & 45.3 & 87.4 &  \multicolumn{1}{c|}{80.3} & 45.5 & \multicolumn{1}{c|}{28.5} & \multicolumn{1}{c|}{63.4/17.2} & \multicolumn{1}{c|}{95.4} & \multicolumn{1}{c}{74.4}\\
BEA-YOLOv3 \cite{qutub2022bea} & 54.8 & 89.3  & \multicolumn{1}{c|}{80.8} & 45.6 & \multicolumn{1}{c|}{91.7} & \multicolumn{1}{c|}{90.5/92.5} & \multicolumn{1}{c|}{33.6} & \multicolumn{1}{c}{18.7}\\
Vanilla-SSD & 62.6 & 88.6  &  \multicolumn{1}{c|}{76.2} & 56.2 & \multicolumn{1}{c|}{35.7} & \multicolumn{1}{c|}{44.3/37.6} & \multicolumn{1}{c|}{96.5} & \multicolumn{1}{c}{55.3}\\
BEA-SSD \cite{qutub2022bea} & 63.1 & 89.6  &  \multicolumn{1}{c|}{74.5} & 54.4 & \multicolumn{1}{c|}{91.6} & \multicolumn{1}{c|}{92.6/90.7} & \multicolumn{1}{c|}{57.6} & \multicolumn{1}{c}{30.6}\\
Vanilla-DINO-DETR & 72.9 & 95.2  &  \multicolumn{1}{c|}{78.2} & 49.2 & \multicolumn{1}{c|}{41.4} & \multicolumn{1}{c|}{42.6/45.0} & \multicolumn{1}{c|}{96.1} & \multicolumn{1}{c}{49.8}\\
BEA-DINO-DETR & 73.8 & 95.8 &  \multicolumn{1}{c|}{79.1} & 52.0 & \multicolumn{1}{c|}{92.4} & \multicolumn{1}{c|}{93.1/94.4} & \multicolumn{1}{c|}{15.1} & \multicolumn{1}{c}{9.6} \\ \hline
\textbf{DBEA-DINO-DETR (ours)}& \textbf{74.6} & \textbf{95.8} &  \multicolumn{1}{c|}{\textbf{79.6}} & 54.2 & \multicolumn{1}{c|}{\textbf{98.3}} & \multicolumn{1}{c|}{\textbf{98.5/98.3}} & \multicolumn{1}{c|}{\textbf{10.3}} & \multicolumn{1}{c}{\textbf{7.6}} \\ \hline \hline
\multicolumn{2}{l}{\textbf{BDD100K trained}}\\ \hline \hline
Vanilla-YOLOv3 & 25.0 & 53.7  &  \multicolumn{1}{c|}{69.3} & 38.7 & \multicolumn{1}{c|}{22.0} & \multicolumn{1}{c|}{27.4/43.6} & \multicolumn{1}{c|}{98.7} & \multicolumn{1}{c}{40.9}\\
BEA-YOLOv3 \cite{qutub2022bea} & 27.6 & 58.1 & \multicolumn{1}{c|}{69.0} & 36.0 & \multicolumn{1}{c|}{96.5} & \multicolumn{1}{c|}{97.5/91.8} & \multicolumn{1}{c|}{12.3} & \multicolumn{1}{c}{8.5}\\
Vanilla-DINO-DETR & 47.3 & 84.0  &  \multicolumn{1}{c|}{71.1} & 37.4 & \multicolumn{1}{c|}{25.3} & \multicolumn{1}{c|}{36.5/35.9} & \multicolumn{1}{c|}{99.8} & \multicolumn{1}{c}{49.6}\\ \hline
\textbf{DBEA-DINO-DETR (ours)} & \textbf{47.9} & \textbf{84.6}  &  \multicolumn{1}{c|}{\textbf{78.9}} & \textbf{48.3} & \multicolumn{1}{c|}{\textbf{99.6}} & \multicolumn{1}{c|}{\textbf{99.7/99.6}} & \multicolumn{1}{c|}{\textbf{1.2}} & \multicolumn{1}{c}{\textbf{2.2}} \\ \hline
\end{tabular}%
}
\caption{Comparison of the performance of our sample-free method with that of previous state-of-the-art sample-free method. Our method is trained on the KITTI and BDD100K datasets, and the out-of-distribution detection evaluation is conducted on the COCO dataset instances.}
\label{tab:benchmark_results}
\end{table*}

\begin{figure}[t]
\centering
    \includegraphics[width=0.95\columnwidth, scale=0.5]{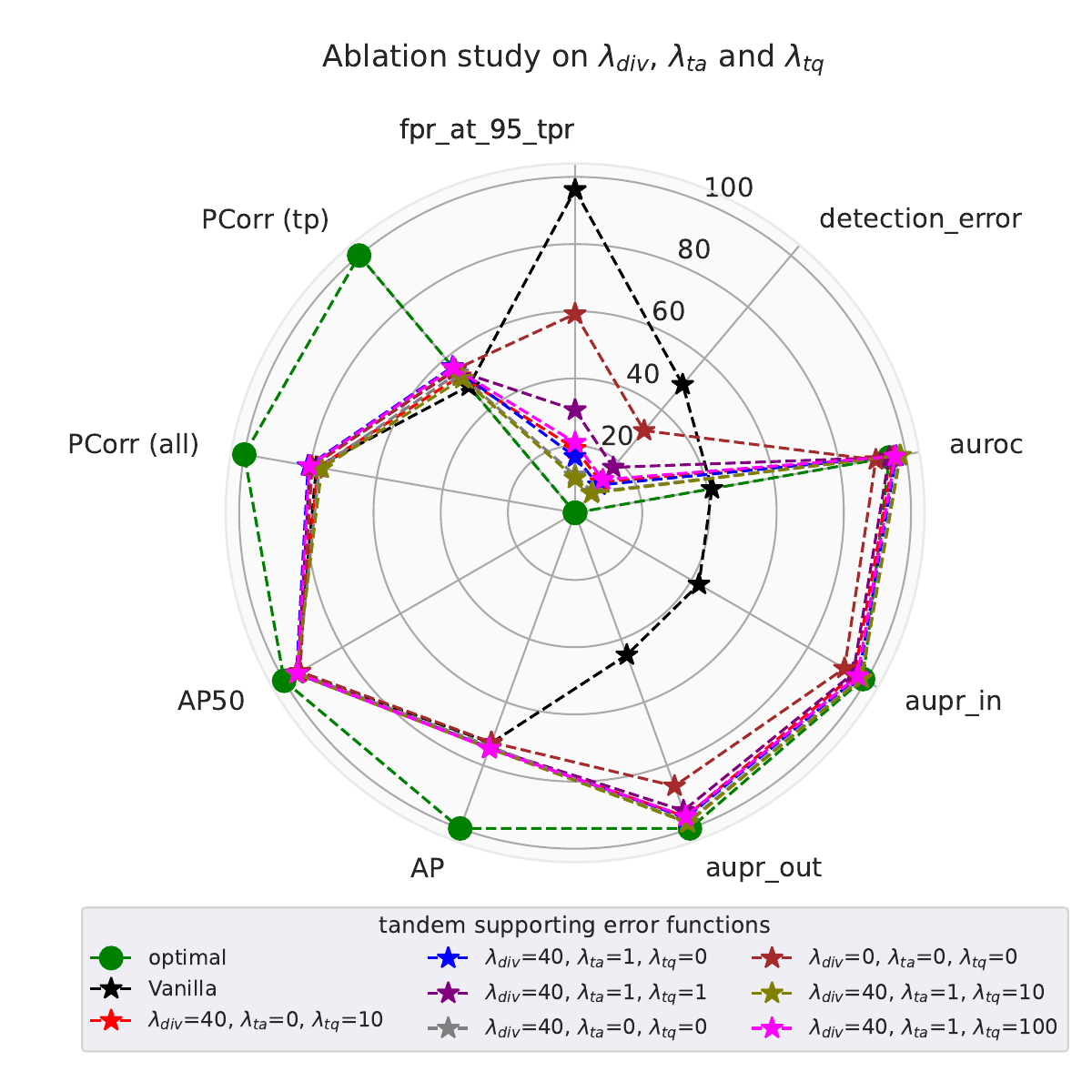}
    \caption{Analysis of the impact of varying parameter $\mathcal{L_{\mbox{tq}}}$ in the ablation study.}
    \label{fig:ablation_all}
\end{figure}

Fig. \ref{fig:ablation_cosin} illustrates the impact of introducing $\mathcal{L_{\mbox{diversity}}}$ with varied $\lambda_{div}$ parameters on the Situation Monitor performance and its OOD detection. The radar plots presented in this study will include the outcomes of a baseline model (the "vanilla model") and reference optimal values against which the metrics will be compared. Without enabling diversity, the performance on the Far-OOD CoCo dataset experiences a significant decline compared to its enabled counterpart. The optimal configuration is identified when the $\lambda_{div}$ parameter is set to $40$, resulting in elevated Average Precision metrics, as well as improved AUROC and AUPR OOD metrics, while concurrently maintaining lower FPR@95 and detection error (DE@95) values. A subsequent increase of $\lambda_{div}$ from $40$ to $80$ brings about enhancements across all metrics, with only a minor impact on detection error and a slight decrease in FPR@95. Hence, choosing $40$ as the parameter proves to be the most effective, achieving a balance across all metrics. The selection of $\lambda_{div}$, incremented by a factor of 2, is influenced by the consideration that both the classification and GIOU loss \cite{giou_loss} components of $\mathcal{L_{\mbox{base}}}$ are scaled by a factor of 2, making factors of 2 more effective to the effectiveness of $\lambda_{div}$.

Similarly, the influence of $\lambda_{tq}$ is illustrated in Fig. \ref{fig:ablation_tq}, with $\lambda_{ta}$ and $\lambda_{div}$ held at 1 and 40, respectively. This specific radar plot highlights that setting $\lambda_{tq}$ to 10 produces the most well-balanced results compared to other factors. While the transformer model exhibits commendable calibration when trained with a $\lambda_{tq}$ factor of 100, there is a slight decline in the Situation Monitor OOD performance.

Fig. \ref{fig:ablation_all} illustrates the results of comprehensive ablation experiments encompassing all three factors: $\lambda_{div}$, $\lambda_{ta}$, and $\lambda_{tq}$. In the absence of diversity and tandem loss, the model exhibits comparable Average Precision (AP) performance but experiences a notable decline in the FPR@95 and DE@95 metrics. Notably, when the model is trained with specific parameters, namely $\lambda_{div}=40$, $\lambda_{ta}=1$, and $\lambda_{tq}=10$, it achieves optimal performance, manifesting in heightened accuracy and superior OOD detection capabilities. The configuration, which results in lower FPR@95 and DE@95 metrics, indicates enhanced OOD detection performance. The improved threshold associated with these parameter values contributes to superior classification, leading to better OOD detection.

\begin{figure}
\centering     
\subfigure[CoCo dataset ($\mathcal{O^\textit{far}}$)]{\label{fig:AUROC_hist_a}\includegraphics[width=0.49\columnwidth]{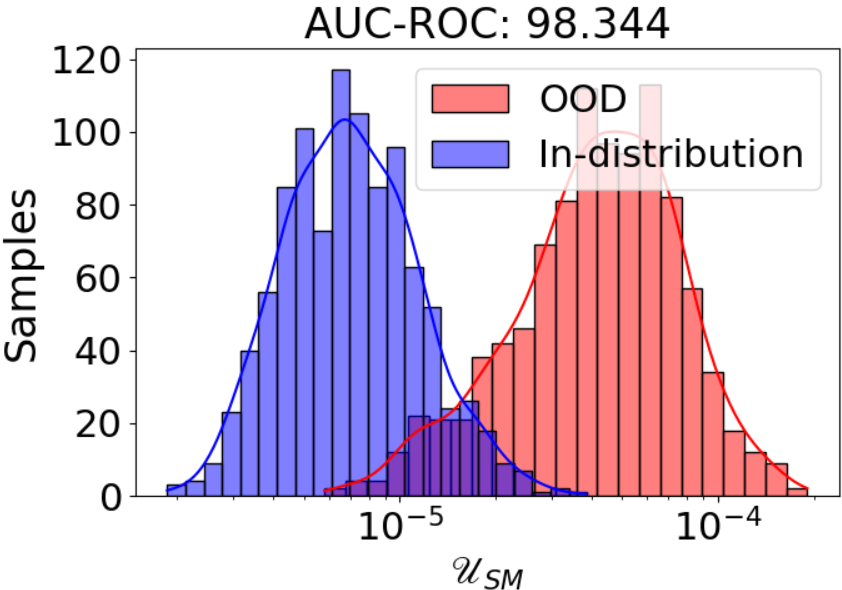}}
\subfigure[BDD100K dataset ($\mathcal{O^\textit{near}}$)]{\label{fig:AUROC_hist_b}\includegraphics[width=0.49\columnwidth]{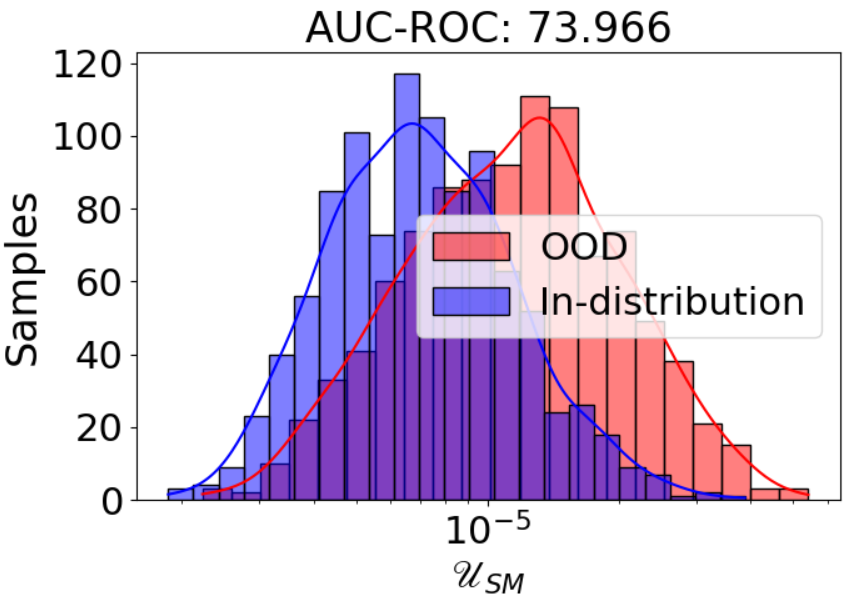}}
\subfigure[Cityscapes dataset ($\mathcal{O^\textit{near}}$)]{\label{fig:AUROC_hist_c}\includegraphics[width=0.49\columnwidth]{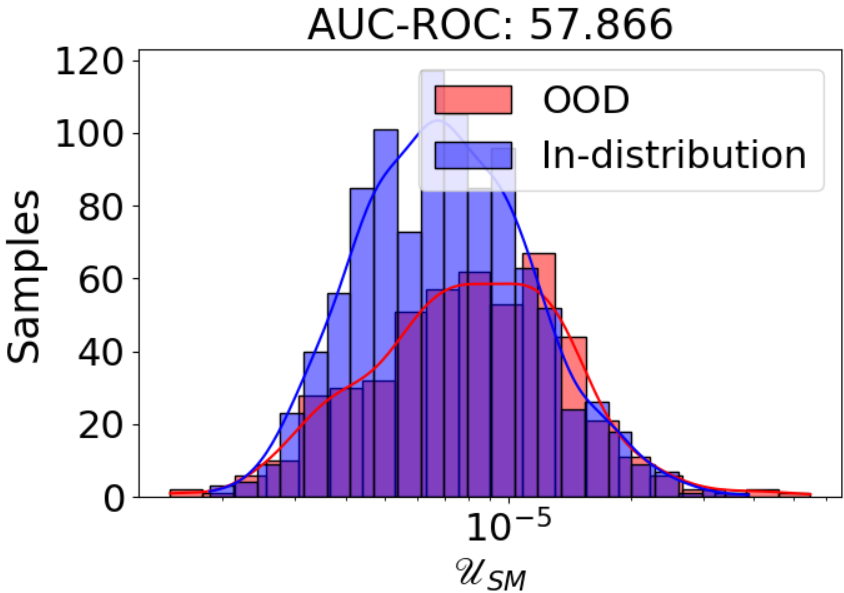}}
\subfigure[Lyft dataset ($\mathcal{O^\textit{near}}$)]{\label{fig:AUROC_hist_d}\includegraphics[width=0.49\columnwidth]{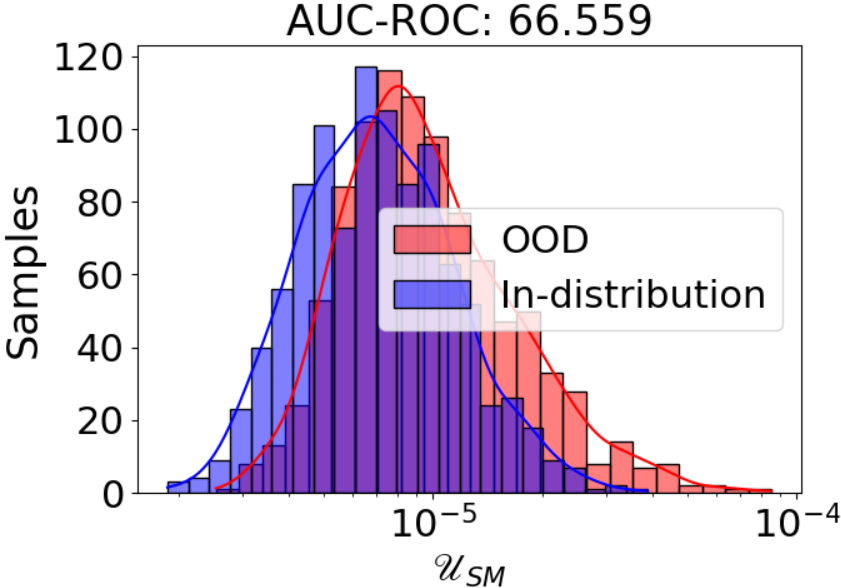}}
\caption{OOD detection performance of KITTI trained DBEA-DINO-DETR model on $\mathcal{O^\textit{far}}$ and $\mathcal{O^\textit{near}}$ datasets. The Situation Monitor of the DBEA-DINO-DETR model can effectively flag Far-OOD situations.}
\label{fig:AUROC_hist}
\end{figure}

\subsection{Benchmark Results}
Tab. \ref{tab:benchmark_results} comprehensively compares baseline models (CNN and transformer-based) with state-of-the-art sample-free OOD detection methods. Using various metrics discussed in Section \ref{sec:metrics}, we assess the Situation Monitor's overall object detection and OOD capabilities integrated into the DBEA-DINO-DETR model. Yolov3, SSD and DINO baseline models are trained using original hyperparameters, and the reported results in Tab. \ref{tab:benchmark_results} reflect their peak performance. Applying our approach to the DINO-DETR model resulted in the DBEA-DINO-DETR model. Training it on datasets KITTI and BDD100K showcases enhanced detection accuracy and improved correlation between predicted confidence scores and intersection over union.
On the BDD100K dataset, our DBEA-based DINO-DETR model demonstrates improved correlation in overall and true positive predictions. Additionally, there is a substantial enhancement in the Out-of-Distribution (OOD) performance for detecting CoCo images. Specifically, the OOD detection performance for CoCo images is higher when the model is trained with BDD100K than the KITTI model. This highlights our approach's effectiveness in advancing object detection and OOD performance, especially on larger datasets.
Illustrated in Fig. \ref{fig:AUROC_hist_a}, the Situation Monitor adeptly identifies $\mathcal{O^\textit{far}}$ OOD samples. Furthermore, the Situation Monitor demonstrates strong generalization to $\mathcal{O^\textit{near}}$ datasets, evident in overlapping histograms of $\mathcal{U}_{SM}$ values Fig. \ref{fig:AUROC_hist_b}, \ref{fig:AUROC_hist_c} and \ref{fig:AUROC_hist_d}. In contrast to the Situation Monitor introduced in this study, the OOD values of BEA-based CNN models \cite{qutub2022bea} tended to miss-classify even $\mathcal{O^\textit{near}}$ as $\mathcal{O^\textit{far}}$ in their OOD detection.
\subsection{Overhead Analysis}
Given the computational intensity of transformers relative to the CNNs, the DBEA adaptation of transformers incurs additional costs due to the duplication of final regression layers. To mitigate this overhead, we limit the feed-forward channels in both the encoder and decoder layers (N) of DBEA-DINO-DETR from 2048 to 1024, offsetting the overhead of duplicating the final regression layers to create the tandem layers. The Vanilla model trained on the KITTI dataset with N=2048 has a size of ~48M, while our DBEA-DINO-DETR is only ~42M, representing a ~14\% reduction compared to the Vanilla model. We also observed that there is little but negligible advantage in maintaining the same number of feed-forward channels of 2048 as the vanilla model. The results of the Situation Monitor presented in this section are based on the DBEA model trained with 1024 feed-forward channels.


\section{Conclusion}

In summary, this paper introduces a Situation Monitor driven by zero-shot learning, which integrates a novel Diversity-based Budding Ensemble Architecture (DBEA) loss function. The incorporation of the DBEA loss function empowers the object detection model to not only identify Far-OOD samples but also to generalize over similar Near-OOD instances effectively. This prevents miss-classification as OOD, enhancing the model's adaptability and robustness.
Through an extensive ablation study with the parameters of DBEA's loss functions, we show significant improvement in detection accuracy and superior OOD detection outcomes compared to baseline models and other existing methods. Additionally, our research demonstrates the scalability of the DBEA-based model, validated through successful training with KITTI and BDD100K datasets. The Situation Monitor is suitable for safety-critical applications, being 14\% less computationally intensive than the baseline DINO-DETR model.



\section*{Acknowledgements}
This work was partially funded by the Federal Ministry for Economic Affairs and Energy of Germany as part of the research project SafeWahr (Grant Number: 19A21026C).

{    
    \small
    \bibliographystyle{ieeenat_fullname}
    \bibliography{bibliography}
}

\end{document}